\documentclass[runningheads]{llncs}

\usepackage{accv}

\usepackage{accvabbrv}

\usepackage{graphicx}
\usepackage{booktabs}
\usepackage{amsmath}

\usepackage[accsupp]{axessibility}  

\usepackage{multirow}

\usepackage{pifont}

\newcommand{\cmark}{\color{blue}{\ding{51}}}
\newcommand{\xmark}{\color{red}{\ding{55}}}

\usepackage[table]{xcolor}

\usepackage{hyperref}

\usepackage{orcidlink}

\newcommand{\ttok}[1]{\texttt{\textless #1\textgreater}}
\begin{document}

\title{TiTok: Audio-Visual LLM \\ for Multi-Segment Temporal Grounding} 

\titlerunning{TiTok}

\author{Eunji Shin\thanks{Equal contribution.
}$^1$, Dahyun Choi$^{\star1}$, Seungyeon Jo$^1$, Yejin Hong$^1$, Jiyoung Lee$^{\dagger1,2}$}

\authorrunning{E.~Shin et al.}

\institute{$^1$Department of Artificial Intelligence, Ewha Womans University \\ 
$^2$Division of Artificial Intelligence \& Software, Ewha Womans University \\[\baselineskip]
\email{\{guma017, cdh030303, serena2140, 2391039, lee.jiyoung\}@ewha.ac.kr}}

\maketitle
{\renewcommand{\thefootnote}{$\dagger$}%
\footnotetext{Corresponding author.}}

\begin{abstract}
Audio-visual multi-segment grounding (AV-MSG) in untri\-mmed videos, reasoning over audio-visual evidence and predicting multiple segments for a query, is a fundamental problem but remains challenging.
Visual-only models overlook complementary acoustic cues, while audio-visual models often fail to calibrate the number of events---a phenomenon we refer to as \emph{count miscalibration}.
We present \textbf{TiTok}, an audio-visual large language model (AV-LLM) that localizes an arbitrary number of temporal event segments for each query.
For precise boundary prediction, we introduce the Time Token Interleaving (TTI) method, which explicitly injects special time tokens into the audio-visual stream to align input-side temporal perception with output-side temporal prediction.
We further propose decoupled, multi-segment-oriented rewards for reinforcement learning, consisting of global, local, count, precision, and format rewards, optimized with Group reward-Decoupled Normalization Policy Optimization (GDPO).
To assess the performance on AV-MSG, we establish a new UnAV-100-based evaluation protocol, and propose the CountF1 metric for quantifying count miscalibration that overlap metrics fail to capture.
TiTok reaches 65.7 mIoU and 0.58 CountF1, achieving state-of-the-art performance.
Our code is available at \url{https://github.com/AI-Living-Lab/TiTok}.
\end{abstract}

\section{Introduction}
 
Video Temporal Grounding (VTG)~\cite{wu2025survey} localizes the temporal intervals of an untrimmed video corresponding to a natural-language query. 
Recent VTG systems have increasingly built on Multimodal Large Language Models (MLLMs) for their cross-modal reasoning and zero-shot generalization~\cite{vtimellm, trace, timer1}, yet most operate almost entirely on the visual stream~\cite{vtimellm, timechat, timer1, museg}, treating audio as auxiliary or discarding it.

\begin{figure}[t]
  \centering
  \includegraphics[width=\linewidth]{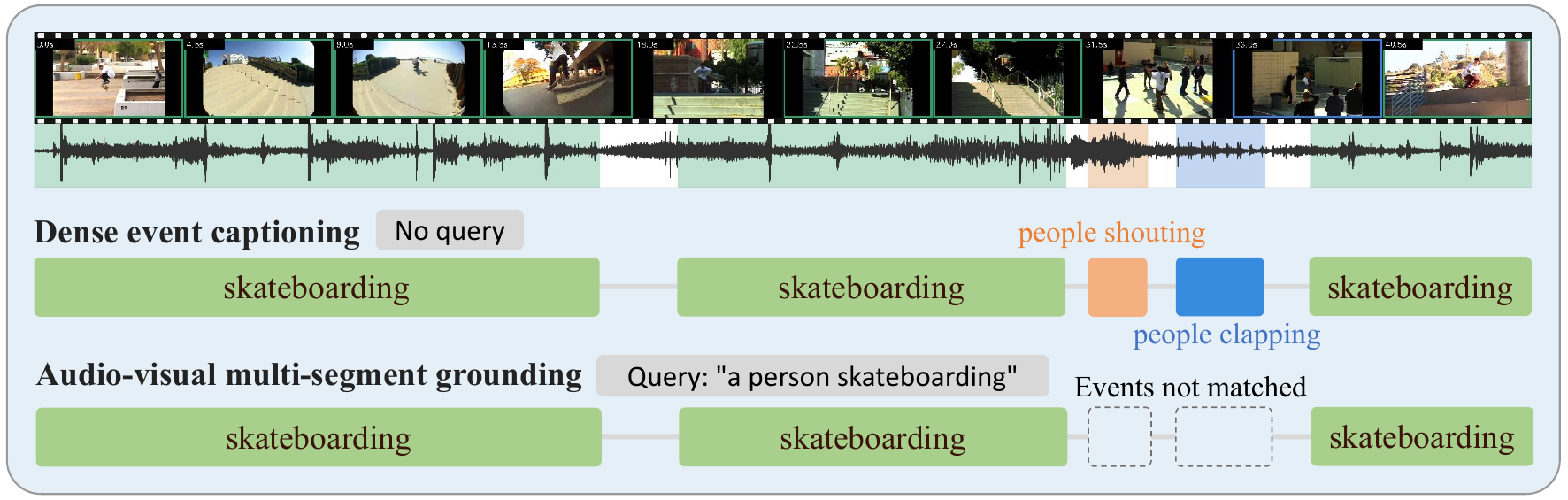}
  \caption{\textbf{Dense event captioning vs. audio-visual multi-segment grounding.} For the same video, dense event captioning (top) describes every event, whereas audio-visual multi-segment grounding (bottom) localizes only the occurrences of the queried event.}
  \label{fig:task}
\end{figure}
 
This visual-centric design overlooks an important regime in which audio is indispensable.
Many real-world events (\eg, a dog barking, a gunshot) are often more clearly delimited temporally by sound than by appearance, and may recur within a single video.
Therefore, given a query event that can appear multiple times, Audio-Visual Multi-Segment Grounding (AV-MSG) identifies all of its temporal occurrences.
Fig.~\ref{fig:task} contrasts AV-MSG with dense event captioning~\cite{krishna2017dense}. The latter localizes all events with timestamps without a query, whereas our task determines which event instances match a given query. 

A growing line of work has extended temporal grounding to the audio-visual setting~\cite{avicuna, archunyuan, chronusomni, ravst, omnivideor1}.
However, accurately counting recurring events remains challenging.
Some models~\cite{archunyuan,chronusomni} have been designed primarily to generate a single event, and other methods~\cite{museg,avicuna} have been shown to collapse occurrences into a single interval or over-segment them.
We attribute these failures to existing objectives that mainly optimize temporal overlap, but do not explicitly penalize errors in event cardinality (\ie, missing or extra event segments).
We argue that this \emph{count miscalibration} is a key bottleneck for robust AV-MSG, a setting existing methods address only partially as described in Tab.~\ref{tab:comparison}.

In this paper, we propose \textbf{TiTok}, which addresses count miscalibration within the AV-LLM framework.
Since existing AV-LLMs~\cite{videosalmonn2, macawllm, pandagpt, videollama, qwen25omni} typically lack an explicit timestamp representation, we resolve this with Time Token Interleaving (TTI), which combines special time tokens with their interleaving into the audio-visual stream. The special time tokens let the model treat time as an explicit, generable quantity distinct from ordinary numerals, and interleaving them marks the absolute time of the audio and visual tokens they precede.

We train TiTok through two stages: a cold-start supervised fine-tuning (SFT) stage that teaches the special time tokens, followed by GDPO-based RL that builds the AV-MSG ability.
However, the reward designs~\cite{timer1,videotgr1} tailored for VTG often fail to account for the correct number of events, since a format and an IoU reward alone do not supervise the event count in the multi-segment setting. 
We therefore design multi-segment-oriented rewards (\ie, global, local, count, precision, and format). 
Combining such heterogeneous rewards into a single scalar suppresses their fine-grained localization signals, so we optimize them with Group reward-Decoupled Normalization Policy Optimization (GDPO)~\cite{gdpo}, which normalizes each reward independently, and further remove the KL penalty and apply asymmetric clipping for freer exploration.
Given that existing benchmarks lack audio-aware multi-segment annotations, we introduce an evaluation protocol on UnAV-100~\cite{unav}. 
TiTok outperforms AVicuna~\cite{avicuna} and ChronusOmni~\cite{chronusomni}, which is fine-tuned on the same training data.
 
Our contributions are summarized as follows:
\begin{itemize}
  \item We propose TiTok, which introduces TTI to interleave absolute timestamps as {special time tokens} into the audio-visual stream, giving the model an explicit, generable time axis for localizing multiple events.
  \item We design decoupled, multi-segment-oriented rewards optimized by a GDPO-based RL pipeline, which preserves the fine-grained localization signals that joint normalization would otherwise wash out.
  \item We establish a UnAV-100-based evaluation protocol for AV-MSG that includes CountF1 to quantify count miscalibration, on which TiTok attains the most accurate AV-MSG results among the compared models.
\end{itemize}

\begin{table}[tb]
  \caption{\textbf{Capability comparison.} Multi-Seg marks whether a method natively emits an arbitrary number of segments; Time Token marks whether it represents time with dedicated tokens (special time tokens) rather than plain-text numbers or percentiles.}
  \label{tab:comparison}
  \centering
  \small
  \setlength{\tabcolsep}{3pt}
  \begin{tabular}{lcccc}
    \toprule
    Method & Audio & Video & Multi-Seg & Time Token \\
    \midrule
    ChronusOmni~\cite{chronusomni}, ARC-Hunyuan~\cite{archunyuan}      & \cmark & \cmark & \xmark & \xmark \\
    TempR1~\cite{tempr1}, MUSEG~\cite{museg}               & \xmark & \cmark & \cmark & \xmark \\
    VTG-LLM~\cite{vtgllm}               & \xmark & \cmark & \xmark & \cmark \\
    AVicuna~\cite{avicuna}              & \cmark & \cmark & \cmark & \xmark \\
    \midrule
    TiTok (Ours)                        & \cmark & \cmark & \cmark & \cmark \\
    \bottomrule
  \end{tabular}
\end{table}

\section{Related Work}

\subsection{Audio-Visual Temporal Grounding with LLMs}
\label{sec:related_av}
With the rapid growth of MLLMs~\cite{qwen25omni,trace}, recent VTG methods~\cite{vtimellm, timechat, trace, timer1, vtgllm} have incorporated {LLMs~\cite{llama,gpt,qwen25}} to leverage their remarkable reasoning capabilities.
Most methods heavily rely on the visual stream to predict a single event timestamp, and are evaluated on benchmarks annotated from visual cues alone~\cite{charades, krishna2017dense}.
Although some studies have extended temporal grounding to the audio-visual setting~\cite{chronusomni, archunyuan, ravst, omnivideor1} or multi-segment prediction~\cite{tempr1, museg}, none addresses both jointly.
While AVicuna~\cite{avicuna} is capable of emitting multiple segments for a query, its predicted segment counts deviate substantially from the ground truth, leaving AV-MSG largely unsolved.


\subsection{Temporal Representation in MLLMs}
\label{sec:related_time}
In MLLMs, time is handled along two axes: (i) on the input side, frame numbers or indices are overlaid on the frames~\cite{numpro, archunyuan}, representations of timestamps are fused with the visual tokens~\cite{timechat}, or time is injected as natural-language text~\cite{timemarker,chronusomni, timelens}.
In AV-LLMs, audio and visual tokens can be interleaved along a shared temporal axis~\cite{videosalmonn2, qwen25omni}.
(ii) On the output side, temporal representations can be divided into relative and absolute time. Relative-time methods represent boundaries using percentiles of video duration~\cite{avicuna}, indices of uniformly sampled frames~\cite{vtimellm}, or dedicated tokens for positions normalized by video length~\cite{seq2time, lita}. 
These representations maintain a fixed output range across videos of different lengths, but the time each unit spans grows with video length, and recovering timestamps in seconds requires conversion. 
In contrast, absolute-time methods express boundaries directly in seconds. Plain-text timestamps~\cite{timemarker,timechat, timer1, chronusomni} can represent times without adding new tokens but share the token space of ordinary numbers. Dedicated absolute-time tokens~\cite{vtgllm} instead use a controlled temporal vocabulary that separates time from ordinary text and reduces quantization error.
However, dedicated time tokens~\cite{seq2time, lita, vtgllm} have been used for visual-only VTG, not interleaved into an audio-visual stream. TiTok addresses this gap with TTI, using the same time tokens at both the input and output.

\subsection{Reinforcement Learning for Temporal Grounding}

Reinforcement learning (RL) has become the dominant approach for strengthening reasoning and temporal understanding in MLLMs~\cite{openaio1, deepseekr1}. 
Many VTG works combine a format reward with an IoU reward for prediction accuracy~\cite{timer1, videotgr1, tvgr1}, while others tune rewards to their task~\cite{museg, segzero, ravst}.
These works typically optimize such distinct rewards together with Group Relative Policy Optimization (GRPO)~\cite{grpo}, leaving its suitability for this multi-reward setting largely unexamined.
Meanwhile, GDPO~\cite{gdpo} reveals that GRPO collapses distinct reward combinations into an identical advantage by normalizing their sum.
This ambiguity weakens the training signal and may lead to suboptimal convergence or early training failure.
Therefore, our method optimizes the designed rewards with GDPO for training.

\section{Method}
\label{sec:methods}

\subsection{Problem Statement}
\label{sec:task}

Given an untrimmed video with visual and audio streams, and a natural-language query describing an event, AV-MSG aims to localize every temporal interval in which the queried event occurs.
Let the video be paired with a query $q$, and let the ground-truth be a set of $N_{\text{gt}}$ segments $\mathcal{G} = \{G_i\}_{i=1}^{N_{\text{gt}}}$, where each $G_i = (s_i, e_i)$ denotes a start time and an end time.
A model must predict a set $\mathcal{P} = \{P_j\}_{j=1}^{N_{\text{pred}}}$ of segments.
The number of predicted segments $N_{\text{pred}}$ is not given in advance but must itself be inferred.
We refer to samples with $N_\text{gt}=1$ as single-segment samples and those with $N_\text{gt}\geq 2$ as multi-segment samples.

\subsection{Time Token Interleaving}
\label{sec:tti}

\begin{figure}[t]
  \centering
  \includegraphics[width=\linewidth]{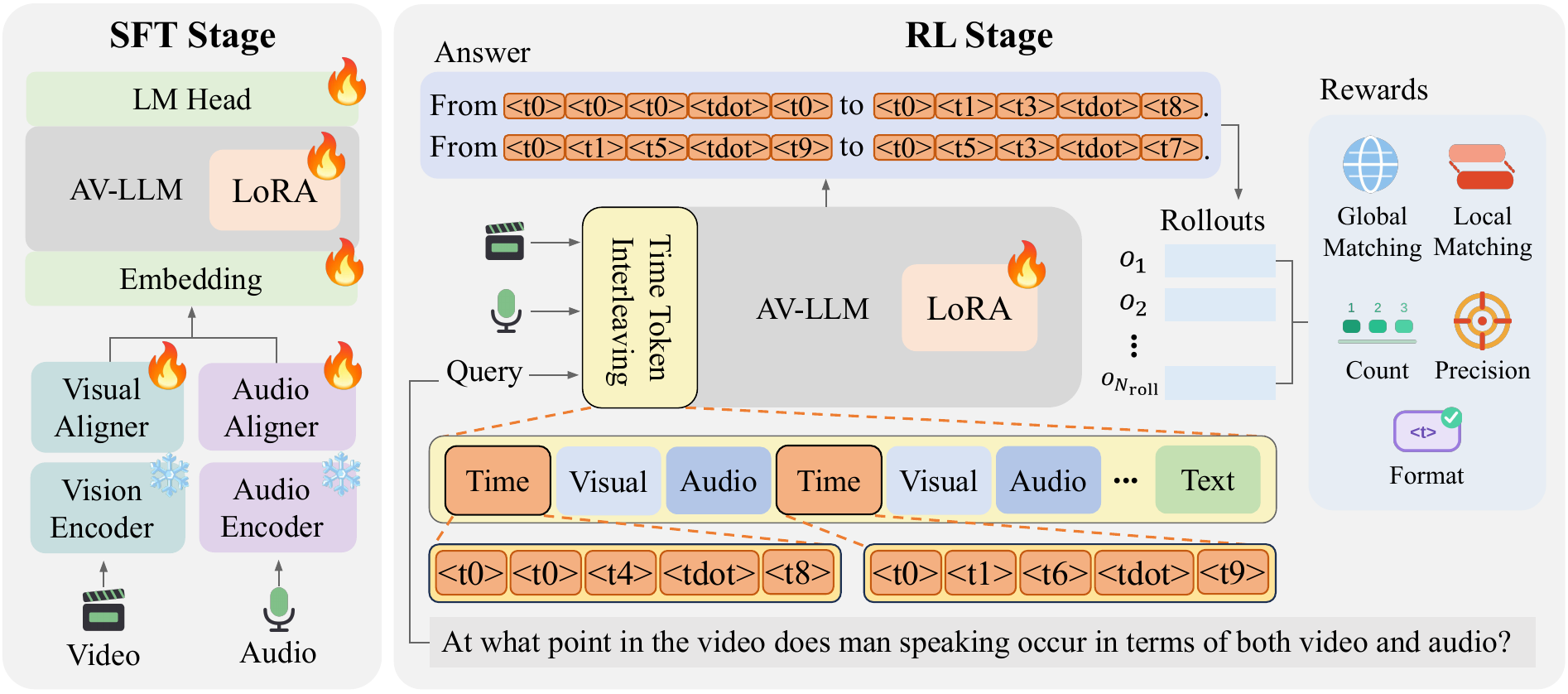}
  \caption{\textbf{The overall pipeline of TiTok.} (Left) Cold-start SFT stage. (Right) GDPO-based RL stage with TTI and the five rewards.}
  \label{fig:main_new}
\end{figure}

We address the lack of absolute time representation in AV-LLMs with TTI.
As shown in Fig.~\ref{fig:main_new}, we interleave audio and visual tokens in temporal order, following video-SALMONN 2 plus~\cite{videosalmonn2}.
Sampling a video into frames thus yields a sequence of audio-visual blocks that each bundle the audio and visual tokens at one timestamp, and into each block we insert that timestamp composed of special time tokens.

Each timestamp is encoded as a composition of tokens rather than a single opaque symbol.
Following VTG-LLM~\cite{vtgllm}, we add ten digit tokens \ttok{t0}~$\sim$~\ttok{t9} and a decimal-point token \ttok{tdot} to the vocabulary and initialize their embeddings and LM head rows from the corresponding digits and decimal point.
A timestamp is written with three zero-padded integer digits and one decimal digit (16.7 seconds $\;\rightarrow\;$ \texttt{\ttok{t0}\ttok{t1}\ttok{t6}\ttok{tdot}\ttok{t7}}), which avoids the concept shift between the numerical and temporal meanings of digits.

TTI uses these special time tokens not only for output timestamps but also in the input stream, where each inserted timestamp marks the absolute time of the audio and visual tokens that follow it.
The model is therefore able to locate the times at which the audio-visual evidence of the queried event appears and disappears, which can enable it to place segment boundaries, distinguish recurring events as separate segments and thus mitigate count miscalibration.
Because the output timestamps share this format, the time perceived at the input and the time predicted at the output are expressed in the same tokens, which supports both boundary prediction and count calibration.

\subsection{Two-Stage Training}
\label{sec:two_stage}

\subsubsection{Stage~1: Cold-Start SFT.}
The special time tokens are new to the pretrained vocabulary, so the backbone rarely generates them in a valid format.
When GDPO is applied directly to the backbone, most responses cannot be parsed into segments and receive zero rewards.
We therefore first perform a short cold-start SFT stage so the model can read and generate the special time tokens for successful rollout generation.

\subsubsection{Stage~2: GDPO-based Reinforcement Learning.}
Once the special time token format is secured, we build AV-MSG ability through RL. 
For a query $q$, we sample $N_\text{roll}$ responses $\{o_1,\dots,o_{N_\text{roll}}\}$ and score each with the $K$ reward terms $\mathbf{R}=\{r^{(1)},\dots,r^{(K)}\}$ of Sec.~\ref{sec:reward}.
Whereas GRPO~\cite{grpo} normalizes the sum of these terms within the group, we adopt GDPO~\cite{gdpo}, which normalizes each term on its own and combines them by weight before normalizing once more across the batch $\mathcal{B}$:

\begin{equation}
\label{eq:gdpo_adv}
\hat{A}_i^{(k)}=\frac{r_i^{(k)}-\operatorname{mean}(\{r_j^{(k)}\}_{j=1}^{N_\text{roll}})}{\operatorname{std}(\{r_j^{(k)}\}_{j=1}^{N_\text{roll}})},
\qquad
\tilde{A}_i=\sum_{k=1}^{K} w^{(k)}\,\hat{A}_i^{(k)},
\end{equation}

\begin{equation}
\label{eq:gdpo_norm}
\hat{A}_i=\frac{\tilde{A}_i-\operatorname{mean}(\{\tilde{A}_j\}_{j\in\mathcal{B}})}{\operatorname{std}(\{\tilde{A}_j\}_{j\in\mathcal{B}})},
\end{equation}
where $\hat{A}_i^{(k)}$ is the group-normalized advantage of reward $k$, $w^{(k)}$ is its combination weight, and $\tilde{A}_i$ is the weighted sum of the per-reward advantages, which Eq.~\eqref{eq:gdpo_norm} normalizes over the batch $\mathcal{B}$ into $\hat{A}_i$.
Normalizing each term separately keeps the fine-grained localization signals from being suppressed by easier-to-learn terms such as the format reward, while batch-level normalization prevents the combined advantage from growing with $K$~\cite{gdpo}.

Using the resulting advantage $\hat{A}_i$, we optimize a clipped surrogate objective over the token-level probability ratio $\rho_{i,t}(\theta)=\frac{\pi_\theta(o_{i,t}\mid q,o_{i,<t})}{\pi_{\theta_\text{old}}(o_{i,t}\mid q,o_{i,<t})}$.
For training stability and exploration, we further adopt two techniques from Decoupled Clip and Dynamic sAmpling Policy Optimization (DAPO)~\cite{dapo}.

\begin{itemize}
    \item \textbf{Clip-Higher.} A tight upper clipping bound suppresses the probability increase of low-probability exploration tokens, causing the policy entropy to collapse~\cite{dapo}. We therefore decouple the clipping range into a lower bound $\epsilon_\text{low}$ and an upper bound $\epsilon_\text{high}$, and set the upper bound larger to leave more room for low-probability exploration tokens to grow.
    \item \textbf{Removing KL divergence.} Since the model must depart substantially from the cold-start SFT policy to acquire the behavior required for \mbox{AV-MSG}, we remove the KL penalty with respect to the reference policy to allow freer exploration~\cite{dapo}.
\end{itemize}

Therefore, the final training objective is

\begin{equation}
\begin{aligned}
\mathcal{J}_\text{TiTok}(\theta)=\mathbb{E}\Big[\frac{1}{N_\text{roll}}\sum_{i=1}^{N_\text{roll}}\frac{1}{|o_i|}\sum_{t=1}^{|o_i|}
\min\Big(&\rho_{i,t}(\theta)\,\hat{A}_i,\ \\
&\operatorname{clip}\big(\rho_{i,t}(\theta),\,1-\epsilon_\text{low},\,1+\epsilon_\text{high}\big)\,\hat{A}_i\Big)\Big]
\end{aligned}
\end{equation}
where $\hat{A}_i$ is the GDPO advantage from Eq.~\eqref{eq:gdpo_norm}.

\subsection{Reward Design}
\label{sec:reward}
We decompose the reward into five objective-specific terms $\mathbf{R}=\{r_\text{global},\,r_\text{local},\\ \,r_\text{count},\,r_\text{prec},\,r_\text{format}\}$, each designed to lie in $[0,1]$.
Following the default of \\ GDPO~\cite{gdpo}, we weight them equally ($w^{(k)}=1$ in Eq.~\eqref{eq:gdpo_adv}) so that each independently normalized reward contributes at full strength without manual tuning.

\subsubsection{Global Reward.}
This evaluates the overall agreement between the predicted and ground-truth segments.
We merge overlapping predicted segments into a union $\widetilde{\mathcal{P}}=\{\widetilde{P}_m\}_{m=1}^{M}$ and compute its temporal precision $\mathrm{Prec}$ and recall $\mathrm{Rec}$ against the ground truth, using their F1 as the reward:

\begin{equation}
\label{eq:global}
\begin{gathered}
\mathrm{Prec}=\frac{\sum_{i}\sum_{m}|G_i\cap\widetilde{P}_m|}{\sum_{m}|\widetilde{P}_m|},
\qquad
\mathrm{Rec}=\frac{\sum_{i}\sum_{m}|G_i\cap\widetilde{P}_m|}{\sum_{i}|G_i|},\\
\qquad
r_\text{global}=\frac{2\,\mathrm{Prec}\cdot\mathrm{Rec}}{\mathrm{Prec}+\mathrm{Rec}}.
\end{gathered}
\end{equation}

\subsubsection{Local Reward.}
While the global reward looks at agreement over the entire event region, the local reward looks at how accurately each individual segment is placed.
We process the predicted segments in their output order and pair each with the unpaired ground-truth segment of the highest Generalized IoU (GIoU)~\cite{giou}.
GIoU provides a non-zero signal even when two segments do not overlap, and following MUSEG~\cite{museg}, we rescale it to $[0,1]$:
\begin{equation}
\operatorname{GIoU}(G_i,P_j)=\frac{1}{2}\left(1+\frac{|G_i\cap P_j|}{|G_i\cup P_j|}-\frac{|C\setminus(G_i\cup P_j)|}{|C|}\right),
\end{equation}
where $C$ is the smallest interval containing both $G_i$ and $P_j$.
Denoting the set of paired predictions by $\mathcal{S}$ and the resulting pairing by $\sigma$, we average GIoU with $\max(N_{\text{gt}},N_{\text{pred}})$ as the denominator so that any count mismatch is reflected:
\begin{equation}
\label{eq:local}
r_\text{local}=\frac{1}{\max(N_{\text{gt}},N_{\text{pred}})}
\sum_{j\in\mathcal{S}}\operatorname{GIoU}\!\left(G_{\sigma(j)},\,P_j\right),
\end{equation}
where the number of paired segments is $\min(N_{\text{gt}},N_{\text{pred}})$.
The unpaired segments thus contribute nothing to the numerator but remain in the denominator.
Pairing each prediction with its best match rather than with the ground-truth segment at the same temporal position avoids the cascading misalignment in which a single misplaced segment throws off all subsequent pairs.

\subsubsection{Count Reward.}
We evaluate how well the number of predicted events matches the number of ground-truth segments:
\begin{equation}
r_{\text{count}}=1-\frac{\left|N_{\text{pred}}-N_{\text{gt}}\right|}{\max\left(N_{\text{pred}}, N_{\text{gt}}\right)}.
\end{equation}
This term penalizes both under- and over-prediction.
However, it can give a prediction that collapses two occurrences into one segment ($N_{\text{gt}}=2$, $N_{\text{pred}}=1$) the same reward as one that predicts two segments for four occurrences ($N_{\text{gt}}=4$, $N_{\text{pred}}=2$), although only the former ignores the recurrence entirely.
We therefore set $r_{\text{count}}=0$ when $N_{\text{gt}}\geq 2$ and $N_{\text{pred}}=1$ to strongly penalize collapsing several events into a single segment.

\subsubsection{Precision Reward.}
In Eq.~\eqref{eq:global}, $r_\text{global}$ combines precision with recall through F1, and since predicting more time can only increase recall, inflating boundaries is penalized there only partially.
We therefore keep precision as a separate term for precision reward, $r_\text{prec}=\mathrm{Prec}$, so that GDPO assigns it an independently normalized advantage.
Whereas the count reward keeps the number of segments correct, the precision reward keeps them tightly placed on the ground truth, discouraging the shortcut of matching the count while inflating boundaries. 

\subsubsection{Format Reward.}
This evaluates whether the model outputs the correct special time tokens.
A reward of $1$ is assigned only when the entire response consists of well-formed segments of the form \texttt{From \ttok{tX}\ttok{tX}\ttok{tX}\ttok{tdot}\ttok{tX} to \ttok{tX}\ttok{tX}\ttok{tX}\ttok{tdot}\ttok{tX}.}, and $0$ otherwise:
\begin{equation}
r_{\text{format}}(o_{i})=
\begin{cases}
1, & \text{if } o_{i} \text{ has the correct format} \\
0, & \text{otherwise}
\end{cases}
\end{equation}
Because any number of well-formed segments earns $1$, this reward enforces strict formatting without discouraging multi-segment predictions.

\section{Experiments}
\subsection{Experimental Settings}
\label{sec:settings}

\subsubsection{Datasets.}
\begin{figure}[t]
  \centering
  \includegraphics[width=\linewidth]{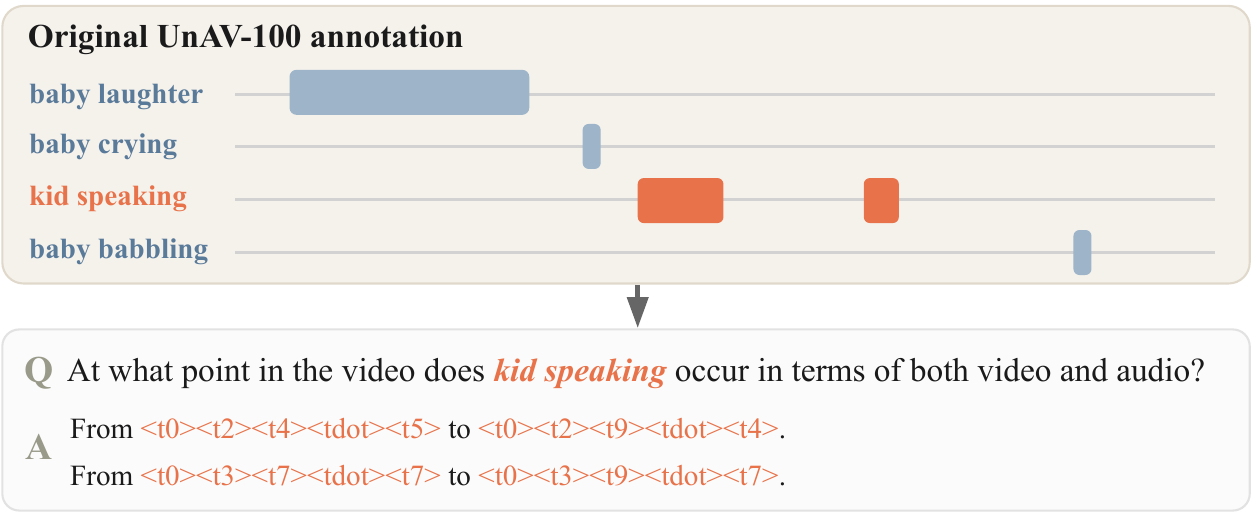}
  \caption{\textbf{UnAV-100 reconstruction.} All intervals of one event category become a single query; a recurring event yields a multi-segment special time token target.}
  \label{fig:unav_convert}
\end{figure}

We build an AV-MSG protocol on UnAV-100~\cite{unav}, a densely annotated audio-visual event dataset spanning 100 categories.
For each video we group annotated intervals by event category into one query--interval sample per category: the category name fills a fixed template (``At what point in the video does \emph{$\langle$event$\rangle$} occur in terms of both video and audio?'') and its chronologically sorted intervals form the target as compositional special time tokens.
A category recurring within a video thus yields a multi-segment ground truth (Fig.~\ref{fig:unav_convert}), while a category that occurs once yields a single-segment one.
For training we also include the moment-retrieval dataset Charades-STA~\cite{charades}, which pairs natural-language queries with video-only inputs and a single target segment.
The RL training set contains 10,358 samples: 7,207 from UnAV-100, of which 4,055 are multi-segment, and 3,151 from Charades-STA, giving a single- to multi-segment ratio of about 6:4.
All evaluation is in-domain on the held-out UnAV-100 test split and on the Charades-STA test split for an additional comparison in Tab.~\ref{tab:newcharades}.
 
\subsubsection{Implementation Details.}
We build TiTok on video-SALMONN 2 plus (7B)~\cite{videosalmonn2} as the backbone and initialize the GDPO-based RL stage from the SFT checkpoint.
The SFT stage uses UnAV-100 training samples only and runs for 250 steps with an effective batch size of 8 and a learning rate of $10^{-4}$.
It updates the visual and audio aligners and the special time token rows of the token embedding and the LM head while applying Low-Rank Adaptation (LoRA)~\cite{lora} ($r$=16, $\alpha$=16) to the LLM, and the vision and audio encoders stay frozen.
In the RL stage we freeze all of these modules and adapt only the LLM through LoRA ($r{=}32$, $\alpha{=}64$) for 2{,}000 steps with an effective batch size of 4 and a learning rate of 10$^{-5}$.
For each prompt we sample $N_\text{roll}$=8 responses with a maximum output length of 256, and perform $\mu$=2 policy update iterations per rollout batch with Clip-Higher ($\epsilon_{\text{low}}$=0.2, $\epsilon_{\text{high}}$=0.28).

\subsubsection{Metrics.}
\label{sec:metrics}
We report \textbf{mIoU}, the per-sample temporal IoU between predicted and ground-truth segment unions, averaged over samples.
As over-prediction inflates recall when the predicted count is unconstrained, we also report \textbf{F1@$k$}~\cite{museg} where $R_k$ ($P_k$) is the fraction of ground-truth (predicted) segments in the test set whose best-overlapping counterpart has a temporal IoU of at least $k$:
\begin{equation}
\text{F1@}k = \frac{2\,P_k \cdot R_k}{P_k + R_k}, \qquad k\in\{0.3,0.5,0.7\}.
\end{equation}
Since overlap metrics do not directly measure the number of segments, we finally introduce \textbf{CountF1} as the harmonic mean of two count-calibration accuracies for opposite failure modes. The under-segmentation accuracy (USA) on multi-segment samples ($S_{\text{multi}}\!=\!\{i:N_{\text{gt}}^{(i)}\!\ge\!2\}$) is high when the model does not predict too few segments, and the over-segmentation accuracy (OSA) on single-segment samples ($S_{\text{single}}\!=\!\{i:N_{\text{gt}}^{(i)}\!=\!1\}$) is high when the model does not split one segment into spurious extras.
For USA, we first compute the count coverage ratio (CR) as the average closeness of the predicted count to the ground-truth count over multi-segment samples.
We rescale CR against a baseline $b$, the expected CR of a trivial model that always predicts a single segment.
Since $b$ depends only on the annotations and not on any evaluated model, it can be recomputed for any benchmark ($b=0.397$ on UnAV-100).
Subtracting $b$ from CR and rescaling to $[0,1]$ yields USA.
For single-segment samples the ground-truth count is fixed at one, so we define OSA directly as the fraction of samples predicted with exactly one segment:
\begin{equation}
\text{CR}=\frac{1}{|S_{\text{multi}}|}\sum_{i\in S_{\text{multi}}}
\frac{\min\!\big(N_{\text{gt}}^{(i)},N_{\text{pred}}^{(i)}\big)}
     {\max\!\big(N_{\text{gt}}^{(i)},N_{\text{pred}}^{(i)}\big)},
\qquad
b=\frac{1}{|S_{\text{multi}}|}\sum_{i\in S_{\text{multi}}}\frac{1}{N_{\text{gt}}^{(i)}}.
\end{equation}
\begin{equation}
\text{USA}=\max\!\Big(0,\,\frac{\text{CR}-b}{1-b}\Big),
\qquad
\text{OSA}=\frac{1}{|S_{\text{single}}|}\sum_{i\in S_{\text{single}}}\mathbf{1}\!\left[N_{\text{pred}}^{(i)}= 1\right].
\end{equation}
\begin{equation}
\text{CountF1}=\frac{2\,\text{USA}\cdot\text{OSA}}{\text{USA}+\text{OSA}},
\end{equation}
where CountF1 is set to 0 when both USA and OSA are 0.
CountF1 lies in [0,1] and is 1 only when a model neither collapses recurring events into too few segments nor inflates single events into too many, so higher CountF1 indicates better count calibration.

\subsubsection{Baseline Prompting and Parsing.}
\label{sec:baseline_eval}
All baselines receive the same query template as TiTok, together with examples
of their native timestamp format (\eg, \texttt{second\{X.X\}} for
ChronusOmni), so that each model answers in the form it was trained to produce. Video frames are likewise sampled using the default settings of each official implementation.
Although ChronusOmni is designed for single-event grounding, we find that it can return several segments under our query template, whereas ARC-Hunyuan always returns one.
We convert the responses into segments with format-tolerant rule-based parsers,
which succeed on 98.9\% of responses on average, and score unparseable responses
as empty predictions. 
To verify that parsing does not understate any baseline, we re-parse 100 randomly
sampled responses per model with an independent LLM-based parser
(Qwen2.5-14B-Instruct\cite{qwen25}). The two parsers extract the same number of segments for
98.3\% of responses, and re-scoring changes mIoU by at most 0.87 points,
suggesting that the reported scores reflect model predictions rather than parsing.

\subsection{Quantitative Results}
\label{sec:main_unav}

\begin{table}[tb]
    \caption{\textbf{Results on UnAV-100.} Setting column marks each model as ZS (zero-shot) or FT (fine-tuned); A and V mark audio and video support. $\dagger$ marks results we reproduced from released checkpoints. }
  \label{tab:unav}
  \centering
  \small
  \setlength{\tabcolsep}{3pt}
  \begin{tabular}{lcccccccc}
    \toprule
    Model & Setting & A & V & mIoU & F1@0.3 & F1@0.5 & F1@0.7 & CountF1 \\
    \midrule
    Qwen2.5-Omni$^{\dagger}$   & ZS & \cmark & \cmark & 12.6 & 14.2 & 9.3 & 6.4 & 0.07 \\
    ARC-Hunyuan$^{\dagger}$    & ZS & \cmark & \cmark & 45.8 & 43.6 & 33.3 & 26.1 & 0.00 \\
    MUSEG$^{\dagger}$          & ZS & \xmark & \cmark & 49.1 & 51.5 & 36.8 & 23.0 & 0.22 \\
    ChronusOmni$^{\dagger}$    & ZS & \cmark & \cmark & 60.9 & 52.7 & 34.8 & 21.8 & 0.39 \\
    \midrule
    AVicuna$^{\dagger}$        & FT & \cmark & \cmark & 50.4 & 31.5 & 23.8 & 18.6 & 0.08 \\
    Crab$^{+}$$^{\dagger}$     & FT & \cmark & \cmark & 50.7 & 48.4 & 36.7 & 27.7 & 0.24 \\
    ChronusOmni                & FT & \cmark & \cmark & 63.4 & 57.5 & 41.8 & 27.9 & 0.50 \\
    TiTok (Ours)               & FT & \cmark & \cmark & \textbf{65.7} & \textbf{63.8} & \textbf{50.3} & \textbf{38.6} & \textbf{0.58} \\
    \bottomrule
  \end{tabular}
  \vspace{1em}
  \centering
\caption{\textbf{Results on Charades-STA.} Both models are fine-tuned on Charades-STA.}
\label{tab:newcharades}
\small
\setlength{\tabcolsep}{6pt}
\begin{tabular}{lccccc}
\toprule
Model & mIoU & F1@0.3 & F1@0.5 & F1@0.7 & OSA \\
\midrule
ChronusOmni  & 48.6 & 74.7 & 50.1 & 24.8 & 0.98 \\
TiTok        & \textbf{56.3} & \textbf{80.3} & \textbf{66.1} & \textbf{39.7} & \textbf{1.00} \\
\bottomrule
\end{tabular}
\end{table}

\subsubsection{Performance on UnAV-100.}
Among the compared methods, AVicuna and Crab$^{+}$~\cite{crab} are released AV-LLMs trained on UnAV-100.
To enable a fairer in-domain comparison, we also fine-tune ChronusOmni, the strongest zero-shot baseline in mIoU and CountF1, with LoRA on the same UnAV-100 and Charades-STA training data as TiTok (Sec.~\ref{sec:settings}).
As shown in Tab.~\ref{tab:unav}, TiTok achieves the best results on every metric among the compared models.
MUSEG, AVicuna, and Crab$^{+}$ support multi-segment prediction but show lower CountF1 and overlap scores across the board. 
Against the fine-tuned ChronusOmni, TiTok attains higher mIoU and outperforms it by a wider margin in F1 and CountF1.
As mIoU compares the unions of predicted and ground-truth segments, a prediction that merges nearby occurrences into one segment can still obtain a high mIoU.
F1@$k$ and CountF1 instead require each occurrence to be predicted as a separate segment, so the wider gap on these metrics indicates that TiTok separates recurring events more accurately rather than covering the same region more tightly.

\subsubsection{Performance on Charades-STA.}
To examine whether TiTok tends to over-segment in single-segment grounding, we additionally evaluate it on Charades-STA against the fine-tuned ChronusOmni used in the preceding UnAV-100 evaluation.
As shown in Tab.~\ref{tab:newcharades}, neither model tends to split a single event into multiple segments, and TiTok outperforms ChronusOmni on every overlap metric by a much wider margin at the stricter F1 thresholds.
TiTok thus remains effective in single-segment visual-only grounding even though it is trained to emit multiple segments.

\subsection{Qualitative Results}
\label{sec:qualitative}

\begin{figure}[t]
  \centering
  \includegraphics[width=\linewidth]{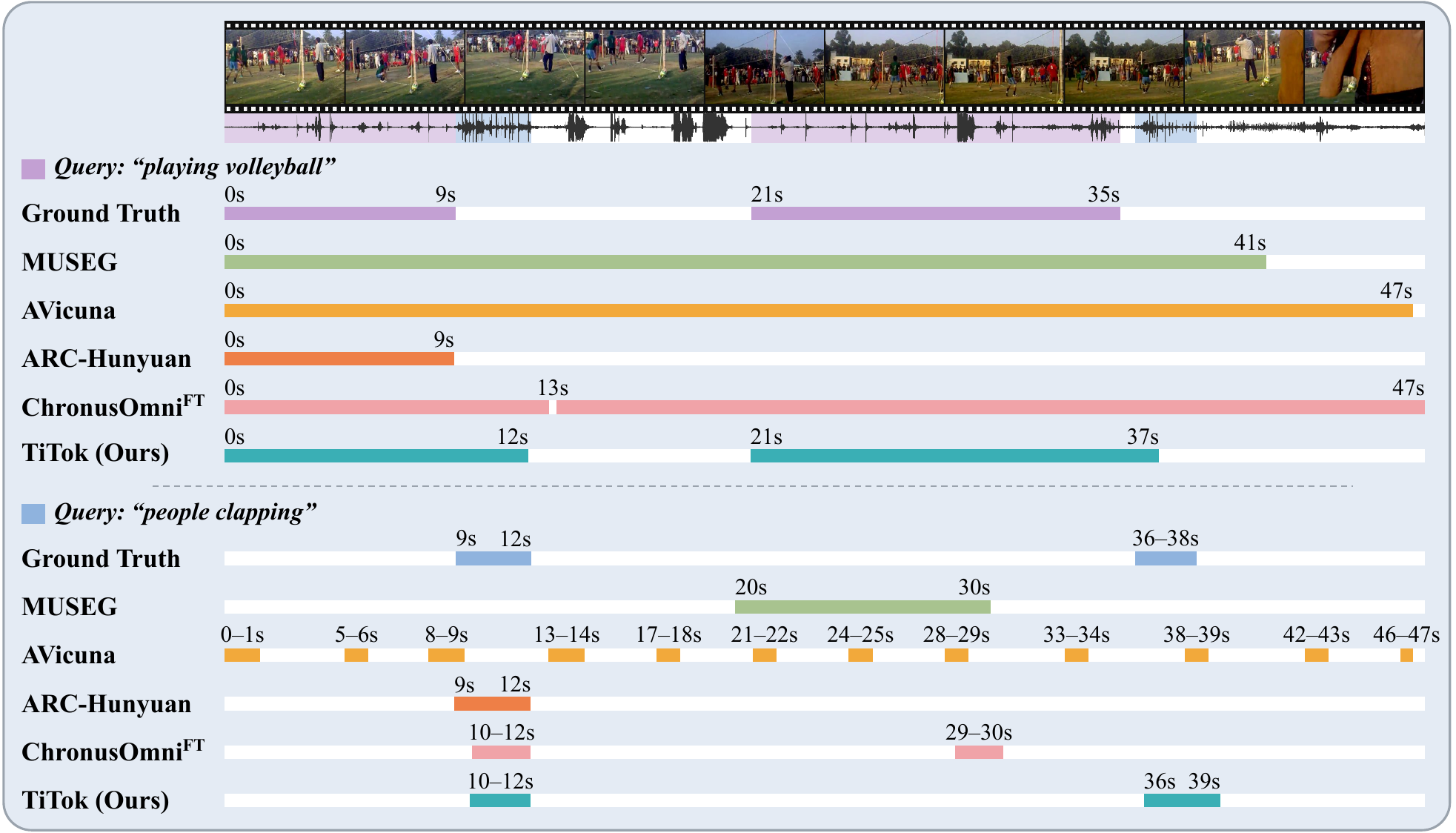}
  \caption{\textbf{Qualitative comparison on UnAV-100.} Two queries on the same clip: ``playing volleyball'' (top) and ``people clapping'' (bottom).}
  \label{fig:qualitative}
\end{figure}

Fig.~\ref{fig:qualitative} illustrates count miscalibration on a clip with two recurring queries: \textit{playing volleyball} is defined mainly by visual cues while \textit{people clapping} is barely visible but clearly audible.
Although MUSEG supports multi-segment prediction, its visual-only input provides almost no cue for \textit{people clapping}, so it localizes the event to the middle of the clip as the most plausible position.
Yet audio input alone does not prevent count miscalibration, as AVicuna collapses the two occurrences of \textit{playing volleyball} into a single interval and over-segments \textit{people clapping} while ARC-Hunyuan localizes only the first occurrence of each query.
The fine-tuned ChronusOmni predicts the correct count for both queries, but its two segments for \textit{playing volleyball} fill the gap between the occurrences and its second segment for \textit{people clapping} lands well before the actual occurrence.
TiTok yields the segment count and boundaries closest to the ground truth for both queries, localizing the recurrences as distinct segments.

\subsection{Ablation Studies}
\label{sec:ablation}
 
\begin{table}[tb]
  \caption{\textbf{Ablation on UnAV-100.} Each row removes or replaces a component.}
  \label{tab:ablation}
  \centering
  \small
  \setlength{\tabcolsep}{4pt}
  \begin{tabular}{lcccccc}
    \toprule
    Setting & mIoU & F1@0.3 & F1@0.5 & F1@0.7 & CountF1 \\
    \midrule
    TiTok                 & \textbf{65.7} & 63.8 & \textbf{50.3} & \textbf{38.6} & 0.58 \\
    \midrule
    w/o special tokens    & 63.9 & 60.6 & 47.3 & 35.5 & 0.51 \\
    w/o interleaving      & 61.2 & 59.8 & 45.6 & 32.7 & 0.53 \\
    \midrule
    w/o audio             & 57.8 & 53.6 & 40.4 & 29.4 & 0.52 \\
    \midrule
    w/o RL                & 56.8 & 26.9 & 19.1 & 14.4 & 0.16 \\
    w/o $r_\text{prec}$  & 65.5 & 62.9 & 49.6 & 36.7 & 0.59 \\
    w/o $r_\text{count}$  & 63.9 & 64.0 & 50.3 & 36.9 & 0.57 \\
    w/o $r_\text{global}$     & 61.4 & 60.7 & 46.4 & 33.7 & 0.57 \\
    w/ GRPO             & 64.4 & \textbf{64.2} & 50.1 & 36.3 & \textbf{0.62} \\
    \bottomrule
  \end{tabular}
\end{table}
\begin{figure}[t]
  \centering
  \includegraphics[width=\linewidth]{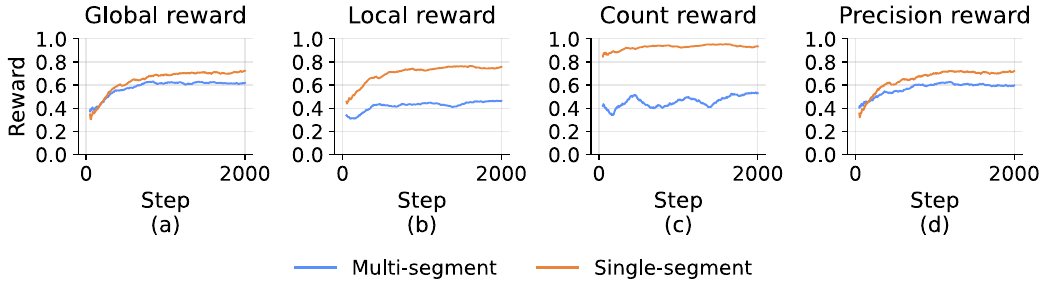}
  \caption{\textbf{Reward curves over GDPO training.}}
  \label{fig:reward_curves}
\end{figure}

\subsubsection{Effect of TTI.}
Removing the special time tokens from the input only (\textit{w/o interleaving}) and replacing them with plain-text timestamps (\textit{w/o special tokens}) both lower every metric but in different patterns.
Compared with \textit{w/o special tokens}, \textit{w/o interleaving} drops more in mIoU and by a wider margin as the F1 threshold becomes stricter, whereas \textit{w/o special tokens} drops more in CountF1.
Interleaving thus contributes more to boundary placement and the special time tokens more to count calibration.
Since \textit{w/o special tokens} still uses the same format at the input and output, its lower CountF1 shows that aligning the two sides alone is not sufficient for count calibration.
TTI therefore requires both components as each addresses a different type of error.

\subsubsection{Effect of Audio.}
Removing audio while keeping TTI and RL (\textit{w/o audio}) lowers every metric, and unlike \textit{w/o interleaving}, its F1 drop does not grow at stricter thresholds.
Audio is therefore needed to detect the occurrences of the queried event and not only to refine their boundaries.
Yet \textit{w/o audio} still exceeds the fine-tuned ChronusOmni in F1@0.7 and CountF1, suggesting that TiTok's advantage in precise localization and count calibration comes from TTI and RL rather than from audio.


\subsubsection{RL Design Choice.}

Each ablated reward term contributes to a different aspect of grounding quality.
Among the reward ablations, removing $r_\text{global}$ (\textit{w/o $r_\text{global}$}) causes the largest drop in mIoU and lowers all F1 scores as well. 
This is because $r_\text{global}$ directly rewards the overall overlap between the predicted and ground-truth segments. 
In contrast, removing $r_\text{count}$ (\textit{w/o $r_\text{count}$}) lowers mIoU and F1@0.7 but leaves F1 at the loose thresholds nearly unchanged because F1 only checks whether each prediction's IoU exceeds the threshold.
With CountF1 also nearly unchanged, the model without $r_\text{count}$ thus misaligns segments slightly rather than miscounting them, likely because the local reward still penalizes count mismatch through its denominator.
Conversely, removing $r_\text{prec}$ (\textit{w/o $r_\text{prec}$}) leaves mIoU nearly unchanged but lowers F1 at every threshold, likely because $r_\text{global}$ still rewards precision over segment unions while $r_\text{prec}$ keeps each segment tightly placed.
With the same reward design, replacing GDPO with GRPO (\textit{w/ GRPO}) raises CountF1 but lowers mIoU and F1@0.7, so neither optimizer is consistently better.
Both optimizers still improve every metric over the cold-start SFT model (\textit{w/o RL}) by a large margin, suggesting that the gains of RL with our reward design do not depend on GDPO.

\subsection{Training Dynamics}
Fig.~\ref{fig:reward_curves} tracks the four reward terms over GDPO training.
The global and precision rewards rise sharply in the early steps and then plateau for both single- and multi-segment samples, so the model learns to place its segments on the ground-truth region early in training.
For the local reward, the single-segment curve sits above the multi-segment one, though the latter also improves slightly over training.
This gap follows from Sec.~\ref{sec:reward}, where multi-segment matching is penalized for any count mismatch (Eq.~\ref{eq:local}) while a single target is easier to cover.
The single-segment count reward stays near its ceiling from the start, while the multi-segment count reward begins well below it and fluctuates during training but trends upward even after the global and precision rewards have plateaued, suggesting that RL continues to reduce count miscalibration on recurring events.

\section{Conclusion}


We propose TiTok, which addresses the count miscalibration problem in AV-LLMs for precise AV-MSG.
It combines TTI with multi-segment-oriented rewards optimized with GDPO.
To make this problem measurable, we also introduce a UnAV-100-based evaluation protocol and CountF1, whose baseline depends only on the annotations and can thus be recomputed for other benchmarks.
On this protocol, TiTok achieves the highest overlap and CountF1 scores among the compared baselines.
These results suggest that reliable AV-MSG benefits from both an explicit representation of time and an explicit objective for the event count.

\subsubsection*{Acknowledgements.}
This research was supported by Global - Learning \& Academic research institution for Master's·PhD students, and Postdocs(G-LAMP) Program of the National Research Foundation of Korea(NRF) grant funded by the Ministry of Education(No. RS-2025-25442252).

%
%
\bibliographystyle{splncs04}
\bibliography{refs}
\end{document}